%% file: main.tex
\documentclass[manuscript,screen]{acmart}

\usepackage[utf8]{inputenc}
\usepackage{amsmath}
\usepackage{amsfonts}
\usepackage{algorithm}
\usepackage{algorithmic}
\usepackage{bigstrut}
\usepackage{multirow,float,tabu,subcaption,enumitem}
\newfloat{algorithm}{t}{lop}
\usepackage{tikz}

\newcommand{\ADER}{\emph{ADER}}
\newcommand{\finetune}{\emph{Finetune}}
\newcommand{\joint}{\emph{Joint}}
\newcommand{\ewc}{\emph{EWC}}
\newcommand{\dropout}{\emph{Dropout}}

\setcopyright{acmcopyright}
\copyrightyear{2020}
\acmYear{2020}
\acmDOI{10.1145/1122445.1122456}

\acmConference[RecSys 2020]{RecSys 2020: ACM Recommender Systems Conference}{Sept. 22--26, 2020}{Virtual Event, Brazil}
\acmBooktitle{The 14th ACM Recommender Systems Conference,
  Sept. 22--26, 2020, Virtual Event, Brazil}
\acmPrice{15.00}
\acmISBN{978-1-4503-XXXX-X/18/06}

\begin{document}

\title[\ADER\ Towards Continual Learning for Session-based Recommendation]{\ADER: Adaptively Distilled Exemplar Replay Towards Continual Learning for Session-based Recommendation}

\author{Fei Mi}
\author{Xiaoyu Lin}
\author{Boi Faltings}
\affiliation
{
\institution{Artificial Intelligence Laboratory, Swiss Federal Institute of Technology Lausanne (EPFL)} 
\city{Lausanne}
 \country{Switzerland} 
 }
\email{firstname.lastname@epfl.ch}

%% article
\begin{abstract}

Session-based recommendation has received growing attention recently due
to the increasing privacy concern.
Despite the recent success of neural session-based recommenders, they are typically developed in an offline manner using a static dataset.
However, recommendation requires continual adaptation to take into
account new and obsolete items and users, and requires ``continual learning'' in real-life applications.
In this case, the recommender is updated continually and periodically
with new data that arrives in each update cycle, and the updated model needs to
provide recommendations for user activities before the next model update.
A major challenge for continual learning with neural models is
catastrophic forgetting, in which a continually trained model forgets
user preference patterns it has learned before.
To deal with this challenge, we propose a method called Adaptively
Distilled Exemplar Replay (\ADER) by periodically replaying previous
training samples (i.e., exemplars) to the current model with an adaptive
distillation loss.
Experiments are conducted based on the state-of-the-art SASRec model
using two widely used datasets to benchmark \ADER\ with several
well-known continual learning techniques. We empirically demonstrate
that \ADER\ consistently outperforms other baselines, and it even outperforms the method using all historical data at every update cycle. This result
reveals that \ADER\ is a promising solution to mitigate the catastrophic
forgetting issue towards building more realistic and scalable
session-based recommenders.

\end{abstract}



% \keywords{datasets, neural networks, gaze detection, text tagging}

\maketitle

\input{text/intro}

\input{text/related}

\input{text/model}

\input{text/experiment}

% \input{text/analysis}

\input{text/conclusion}

\bibliographystyle{ACM-Reference-Format}
\bibliography{main.bib}
% \printbibliography
\end{document}

%% file: text/intro.tex
\vspace{-0.1in}
\section{Introduction}

Due to new privacy regulations that prohibit building user
preference models from historical user data, utilizing anonymous short-term interaction data within a browser session becomes popular. Session-based Recommendation (SR) is
therefore increasingly used in real-life online
systems, such as E-commerce and social media. The goal of SR is to make recommendations based
on user behavior obtained in short web browser sessions, and
the task is to predict the user's next actions, such as clicks, views, and even purchases,
based on previous activities in the same session.

Despite the recent success of various neural approaches \cite{hidasi2015session,li2017neural,liu2018stamp,kang2018self}, they are developed in an offline manner, in which the recommender is trained on a very large static training set and evaluated on a very restrictive testing set in a \textit{one-time} process. 
However, this setup does not reflect the realistic use cases of online recommendation systems. 
In reality, a recommender needs to be periodically updated with new data steaming in, and the updated model is supposed to provide recommendations for user activities before the next update. In this paper, we propose a continual learning setup to consider such realistic recommendation scenarios.

The major challenge of continual learning is \textit{catastrophic forgetting} \cite{mccloskey1989catastrophic,french1999catastrophic}. That is, a neural model updated on new data distributions tends to forget old distributions it has learned before. A naive solution is to retrain the model using all historical data every time.  However, it suffers from severe computation and storage overhead in large-scale recommendation applications.

To this end, we propose to store a small set of representative sequences from previous data, namely \textit{exemplars}, and replay them each time when the recommendation model needs to be trained on new data.  Methods using exemplars have shown great success in different continual learning \cite{rebuffi2017icarl,castro2018end} and reinforcement learning \cite{schaul2015prioritized,andrychowicz2017hindsight} tasks. In this paper, we propose to select representative exemplars of an item using an \textit{herding} technique \cite{welling2009herding,rebuffi2017icarl}, and its exemplar size is proportional to the item frequency in the near past.
To enforce a stronger constraint on not forgetting previous user preferences, we propose a regularization method based on the well-known knowledge distillation technique \cite{hinton2015distilling}. 
We propose to apply a distillation loss on the selected exemplars to preserve the model's knowledge. The distillation loss is further adaptively interpolated with the regular cross-entropy loss on the new data by considering the difference between new data and old ones to flexibly deal with different new data distributions.

Altogether, (1) we are the first to study the practical continual learning setting for the session-based recommendation task;  (2) we propose a method called Adaptively Distilled Exemplar Replay (\ADER) for this task, and benchmark it with state-of-the-art continual learning techniques; (3) experiment results on two widely used datasets empirically demonstrate the superior performance of \ADER\ and its ability to mitigate catastrophic forgetting.\footnote{Code is available at: \url{https://github.com/DoubleMuL/ADER}}

%% file: text/related.tex
\vspace{-0.06in}
\section{Related Work}

 \vspace{-0.02in}
\subsection{Session-based Recommendation}

Session-based recommendation (SR) can be formulated as a sequence learning problem to be solved by recurrent neural networks~(RNNs).
      The first work (GRU4Rec, \cite{hidasi2015session}) uses a gated recurrent unit (GRU) to learn session representations from previous clicks.
      Based on GRU4Rec, \cite{hidasi2018recurrent} proposes new ranking losses on relevant sessions, and \cite{tan2016improved} proposes to augment training data.
     Attention operation is first used by NARM \cite{li2017neural} to pay attention to specific parts of the sequence.  
    Base on NARM, \cite{liu2018stamp} proposes STAMP to model users’ general and short-term interests using two separate attention operations, and \cite{ren2018repeatnet} proposes RepeatNet to predict repetitive actions in a session.
    Motivated by the recent success of \textit{Tansformer} \cite{vaswani2017attention} and \textit{BERT} \cite{devlin2018bert} for language model tasks, \cite{kang2018self} proposed SASRec using \textit{Transformer}, and \cite{sun2019bert4rec} proposed BERT4Rec to model bi-directional information.
    Despite the broad exploration and success, the above methods are all studied in a static and offline manner. Recently, the incremental and steaming nature of SR is pointed out by \cite{guo2019streaming,mi2020memory}.
  
    Besides neural approaches, several non-parametric methods have been proposed. 
      \cite{jannach2017recurrent} proposed \texttt{SKNN} to compare the current session with historical sessions in the training data.
      Lately, variations~\cite{ludewig2018evaluation,garg2019sequence} of \texttt{SKNN} have been proposed to consider the position of items in a session or the timestamp of a past session. \cite{garcin2013personalized,mi2016adaptive,mi2017adaptive,mi2018context} applies a non-parametric structure called \textit{context tree}.
      Although these methods can be efficiently updated, the realistic continual learning setting and the corresponding forgetting issue remain to be explored.
      
      \vspace{-0.05in}
      \subsection{Continual Learning}
      
      The major challenge for continual learning is catastrophic forgetting~\cite{mccloskey1989catastrophic,french1999catastrophic}. Methods designed to mitigate catastrophic forgetting fall into three categories: regularization~\cite{li2017learning,kirkpatrick2017overcoming,zenke2017continual}, Exemplar Replay~\cite{rebuffi2017icarl,chaudhry2019continual,castro2018end} and dynamic architectures~\cite{rusu2016progressive,maltoni2019continuous}. Methods using dynamic architectures increase model parameters throughout the training process, which leads to an unfair comparison with other methods. In this work, we focus on the first two categories.

\textbf{Regularization} methods add specific regularization terms to consolidate knowledge learned before. \cite{li2017learning} introduces knowledge distillation~\cite{hinton2015distilling} to penalize model logit change, and it is widely employed by~\cite{rebuffi2017icarl,castro2018end,wu2019large,hou2019learning,zhao2019maintaining}. \cite{kirkpatrick2017overcoming,zenke2017continual,aljundi2018memory} propose to penalize changes on parameters that are crucial to old knowledge according to various importance measures.
\textbf{Exemplar Replay} methods store past samples, a.k.a \textit{exemplars}, and replay them periodically to prevent model forgetting previous knowledge.
Besides selecting exemplars uniformly, \cite{rebuffi2017icarl} incorporates the \textit{Herding} technique~\cite{welling2009herding} to select exemplars, and it soon becomes popular~\cite{castro2018end,wu2019large,hou2019learning,zhao2019maintaining}. 
%\cite{ramalho2019adaptive} proposes to store the most ``surprising'' samples that the model is least confident.

%% file: text/model.tex
\section{Methodology}

In this section, we first introduce some background in Section \ref{subsec:background} and a formulation of the continual learning setup in Section \ref{subsec:formulation}.
In Section \ref{subsec:model}, we propose our method called ``Adaptively Distilled Exemplar Replay'' (\ADER).

\vspace{-0.03in}
\subsection{Background on Neural Session-based Recommenders}
\label{subsec:background}

A user \textit{action} in SR is a click or view on an item, and the task is to predict the next user action based on a sequence of user actions in the current web-browser session.
Existing neural models $f(\theta)$ typically contain two modules: an \textbf{feature extractor} $\phi(\mathbf{x})$ to compute a compact \textit{sequence representation} of the sequence $\mathbf{x}$ of previous user actions, and an \textbf{output layer} $\omega(\phi(\mathbf{x}))$ to predict the next user action. 
Various recurrent neural networks \cite{hidasi2015session,hidasi2018recurrent} and attention mechanisms \cite{li2017neural,liu2018stamp,kang2018self} have been proposed for $\phi$, and the common choices for the output layer $\omega$ is fully-connect layers\cite{hidasi2015session} or bi-linear decoders \cite{li2017neural,kang2018self}.
In this paper, we base our comparison on SASRec \cite{kang2018self}, and we refer readers to model details in the original paper to avoid verbosity. Nevertheless, the techniques proposed and compared in this paper are agnostic to $f(\theta)$, therefore, a more thorough comparison using different $f(\theta)$ are left for interesting future work.

\begin{figure*}[t!]
        \centering
        \includegraphics[width=0.75\textwidth]{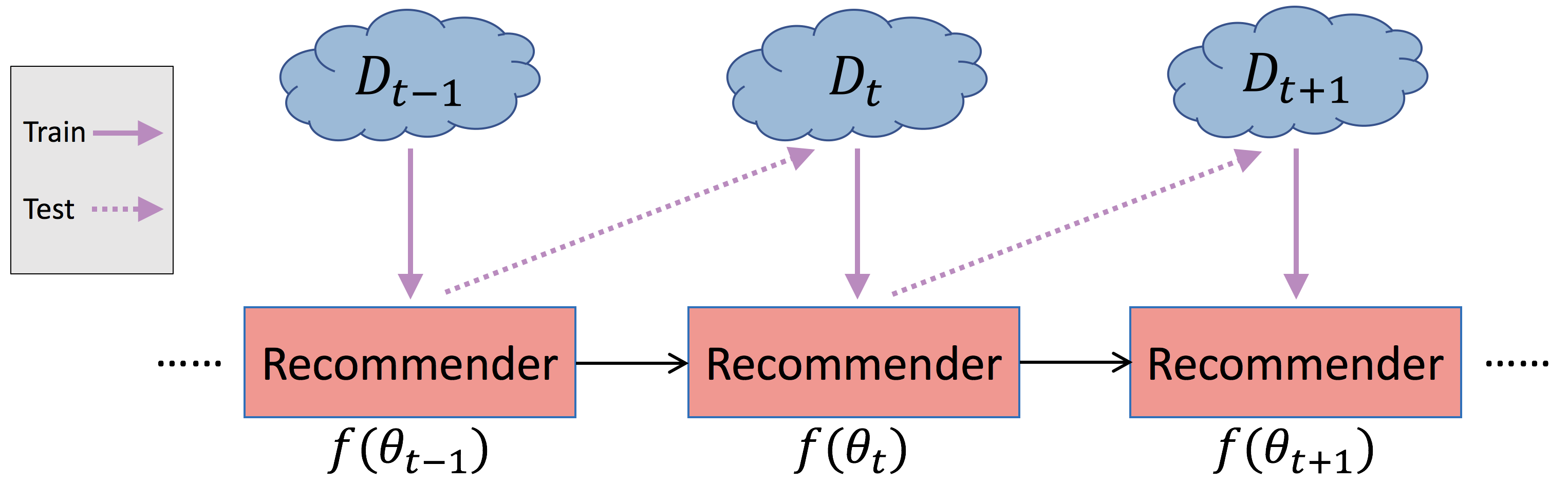} 
 \vspace{-0.1in} \caption{An visualization of the continual learning setup. At each update cycle $t$, the model is trained with data $D_t$, and the updated model $f(\theta_t)$ is evaluated w.r.t. to data $D_{t+1}$ before the next model update.} \vspace{-0.12in}
        \label{fig:formulation}
\end{figure*}    

\vspace{-0.03in}
\subsection{Formulation of Continual Learning for Session-based Recommendation}
\label{subsec:formulation}

In this section, we formulate the continual learning setting for the session-based recommendation task to simulate the realistic use cases of training a recommendation model continually.
%In real-life recommendation scenarios, a model is updated periodically and the updated model needs to perform recommendations for user activities before the next update cycle. 
To be specific, at an update cycle $t$, the recommendation model $f(\theta_{t-1})$ obtained until the last update cycle $t-1$ needs to be updated with new incoming data $D_t$. After $f(\theta_{t-1})$ is trained on $D_t$, the updated model $f(\theta_t)$ is evaluated w.r.t. the incoming data $D_{t+1}$ before the next update cycle $t+1$.
A visualization of the continual learning setup is illustrated in Fig. \ref{fig:formulation}, where a recommendation model is continually trained and tested upon receiving data in sequential update cycles.

\vspace{-0.03in}
\subsection{Proposed Solution: Adaptively Distilled Exemplar Replay (\ADER)}
\label{subsec:model}

        \subsubsection{\textbf{Exemplar Replay}}
%A widely-recognized issue in continual learning is catastrophic forgetting \cite{mccloskey1989catastrophic,french1999catastrophic}, where a model trained on new data distributions tends to forget old distributions it has learned before. 
        To alleviate the widely-recognized catastrophic forgetting issue in continual learning, the model needs to preserve old knowledge it has learned before. 
        To this end, we propose to store past samples,a.k.a \textit{exemplars}, and replay them periodically to preserve previous knowledge.
        To maintain a manageable memory footprint, we only store a fixed total number of exemplars throughout the entire continual learning process. 
        Two decisions need to be made at each cycle $t$: (1). how many exemplars should be stored for each item/label? (2). what is the criterion for selecting exemplars of an item/label?
       
       First, we design the number of exemplars of each appeared item in $I_t$ (i.e. the set of appeared items until cycle $t$) to be proportional to its appearance frequency. In other words, more frequent and popular items contribute a larger portion of selected exemplars to be replayed to the next cycle.
        Suppose we store $N$ exemplars in total, the number of exemplars $m_{t,i}$ at cycle $t$ for a item $i \in I_t$ is:
        \begin{equation}
        m_{t,i} = N \cdot \frac{| \{\mathbf{x},y=i\} \in  D_t \cup E_{t-1}| }{| D_t \cup E_{t-1} |},
        \label{eq:mem}
        \end{equation}
        where the second term is the probability that item $i$ appears in the current update cycle, as well as in the exemplars $E_{t-1}$ we kept from the last cycle.
Therefore, the exemplar sizes of different items to be select in the cycle $t$ can be encoded as a vector $M_t=[m_1, m_2, ..., m_{|I_t|} ]$.
        
        Second, we need to decide which samples to select as exemplars for each item.
 We propose to use a herding technique \cite{welling2009herding,rebuffi2017icarl} to select the most representative sequences of an item in an iterative manner based on the distance to the mean feature vector of the item. 
 %To selected $m_i$ exemplars for item $i$, the herding technique selected sample in an iterative manner until the target number $m_i$ is met. 
 In each iteration, one sample from $D_t \cup E_{t-1}$ that best approximates the average feature vector ($\mu$) over all training examples of this item ($y$) is selected to $E_t$.
 The details are presented in Algorithm \ref{alg:ExemplarSelection}.  

\vspace{-0.05in} 
     \subsubsection{\textbf{Adaptive Distillation on Exemplars}}

The number of exemplars should be reasonably small to reduce memory overhead. As a consequence, the constraint to prevent the recommender forgetting previous user preference patterns is not strong enough.
To enforce a stronger constrain on not forgetting old user preference patterns, we propose to use a knowledge distillation loss \cite{hinton2015distilling} on exemplars to better consolidate old knowledge 

 \input{algorithm/algorithm}
 
At a cycle $t$, the set of exemplars to be replayed is $E_{t-1}$ and the set of items till the last cycle is $I_{t-1}$, the proposed knowledge distillation (KD) loss is written as:
\begin{equation}
L_{KD}(\theta_t) = - \frac{1}{|E_{t-1}|} \sum\nolimits_{(\mathbf{x},y) \in E_{t-1}} \sum\nolimits_{i=1}^{|I_{t-1}|}\hat{p}_i \cdot log(p_i ), 
\label{eq:distill}
\end{equation}
where $[\hat{p}_1, \dots, \hat{p}_{|I_{t-1}|}]$ is predicted distribution (softmax of logits) over $I_{t-1}$ generated by $f(\theta_{t-1})$, and $ [{p}_1, \dots, {p}_{|I_{t-1}|}]$  is the prediction of $f({\theta_t})$ over $I_{t-1}$.
$L_{KD}$ measures the difference between the outputs of the previous model and the current model on exemplars, and the idea is to penalize prediction changes on items in previous update cycles.

L$_{KD}$ defined above is interpolated with a regular cross-entropy (CE) loss computed w.r.t. $D_t$ defined below:
\begin{equation}
L_{CE}(\theta_t) = - \frac{1}{|D_{t}|} \sum\nolimits_{(\mathbf{x},y) \in D_{t}} \sum\nolimits_{i=1}^{|I_{t}|} \delta_{i=y} \cdot log(p_i ), 
\end{equation}        
In practice, the size of incoming data and the number of new items varies in different cycles, therefore, the degree of need to preserve old knowledge varies.
To this end, we propose an adaptive weight $\lambda_t$ to combine $L_{KD}$ with $L_{CE}$:
        \begin{equation}
       L_{ADER} = L_{CE} + \lambda_t \cdot L_{KD},   \quad  \lambda_{t} = \lambda_{base}\sqrt{\frac{|I_{t-1}|}{|I_t|}\cdot\frac{|E_{t-1}|}{|D_{t}|}}
       \label{eq:ader_loss}
        \end{equation}
     In general, $\lambda_t$ increases when the ratio of the number of old items to that of new items increases, and when the ratio of the exemplar size to the current data size increases. The idea is to rely more on L$_{KD}$ when the new cycle contains fewer new items or fewer data to be learned. The overall training procedure of \ADER\ is summarized in Algorithm \ref{alg:UpdateModel}.

%% file: algorithm/algorithm.tex
\begin{figure}[!t]
\begin{minipage}[t]{0.5\textwidth}
    \begin{algorithm}[H]
    \caption{\ADER: ExemplarSelection at cycle $t$}
    \label{alg:ExemplarSelection}
        \begin{algorithmic}
        \REQUIRE{$\mathcal{S}=D_t\cup E_{t-1}$}; $M_t=[m_1, m_2, ..., m_{|I_t|} ]$
        %\REQUIRE{$\mathbf{M_t}=[m_1, m_2, ..., m_{|I_t|} ]$}
        \FOR{$y = 1,...,|I_t|$}
            \item 
            $\mathcal{P}_{y} \gets \{\mathbf{x}: \forall (\mathbf{x}, y) \in \mathcal{S}  \}$
            \item $\mu\gets\frac{1}{|\mathcal{P}_y|}\sum_{\mathbf{x}\in\mathcal{P}_y}\phi(\mathbf{x})$
            \FOR{$k = 1,...,m_y$} 
            \item $\mathbf{x}^k\gets\mathop{\arg\min}_{\mathbf{x}\in\mathcal{P}_y}\|\mu-\frac{1}{k}[\phi(\mathbf{x})+\sum_{j=1}^{k-1}\phi(\mathbf{x}^j )]\|$
            % \item
            % $e_y^k\gets(\mathbf{x}^k, y)$
            \ENDFOR
            \item 
            $E_y \gets \{(\mathbf{x}^1, y),...,(\mathbf{x}^{m_y}, y)\}$ 
            % \item 
            % $E_y \gets \{e_y^1,...,e_y^{m_y}\}$ 
        \ENDFOR
        %\item $E_t \gets \cup_{y=1}^{|I_t|}E_y $
        
        \ENSURE exemplar set $E_t=\cup_{y=1}^{|I_t|}E_y$ 
        
        \end{algorithmic}
    \end{algorithm} \vspace{-0.2in}
\end{minipage}
\hfill
\begin{minipage}[t]{0.46\textwidth}
    \begin{algorithm}[H]
    \caption{\ADER: UpdateModel at cycle $t$}
    \label{alg:UpdateModel}
        \begin{algorithmic}
        \REQUIRE{$D_t, E_{t-1}, I_t, I_{t-1}$} 
        \STATE Initialize $\theta_{t}$ with $\theta_{t-1}$
        \WHILE {$\theta_t$ not converged}
        \STATE Train $\theta_t$ with loss in Eq. (\ref{eq:ader_loss})
        \ENDWHILE
        \STATE Compute $E_t$ using Algorithm \ref{alg:ExemplarSelection} with $\theta_t$ and $M_t$ computed by Eq. (\ref{eq:mem})
        \ENSURE updated $\theta_t$ and new exemplar set $E_t$ 
        \end{algorithmic}
    \end{algorithm}
\end{minipage}
\hfill
\end{figure}

%% file: text/experiment.tex
\section{Experiments}
\label{sec:exp}
  %  In this section, we employ ADER in our experiments. We first introduce the datasets, some state-of-the-art continue learning methods, and the evaluation metrics. Finally, we compare ADER with those methods.
    
    \subsection{Dataset}
    \label{subsec:dataset}
Two widely used dataset are adopted: (1). \textbf{DIGINETICA}: This dataset contains click-streams data
on a e-commerce site over a 5 months, and it is used for
CIKM Cup 2016 (\url{http://cikm2016.cs.iupui.edu/cikm-cup}). (2). \textbf{YOOCHOOSE}: It is another dataset used by RecSys Challenge 2015 (\url{http://2015.recsyschallenge.com/challenge.html}) for predicting click-streams on another e-commerce site over 6 months.
    
   As in \cite{hidasi2015session,li2017neural,liu2018stamp,kang2018self}, we remove sessions of length 1 and items that appear less than 5 times. To simulate the continual learning scenario, we split the model update cycle of DIGINETICA by \textit{weeks} and YOOCHOOSE by \textit{days} as its volume is much larger. Different time spans also resemble model update cycles at different granulates. In total, 16 update cycles are used to continually train the recommender on both datasets. 10\% of the training data of each update cycle is randomly selected as a validation set. Statistics of split datasets are summarized in Table \ref{tab:dataset}. We can see that YOOCHOOSE is less dynamic, indicated by the tiny fraction of actions on new items, that is, old items heavily reappear.

\input{table/dataset}       
    
    \subsection{Evaluation Metrics}
Two commonly used evaluation metrics are used: (1). \textbf{Recall@k}: The ratio when the desired item is among
the top-k recommended items. 
%It can be interpreted as precision \cite{liu2018stamp,wu2019large} because we predict the immediate next event. 
(2). \textbf{MRR@k}: Recall@k does not consider the order of the items recommended, while MRR@k measures the \textit{mean reciprocal ranks} of the desired items in top-k recommended items.
For easier comparison, we reported the mean value of these two metrics averaged over all 16 update cycles.

    \subsection{Baseline Methods}
    \label{subsec:baseline}
    Several widely adopted baselines in continual learning literature are compared:
    %To show the improvement of the proposed method, we introduce three types of continue learning baselines. Tow use regularization (Dropout, EWC), one uses exemplar replay (Exemplar) and final jointing training as an upper bound.
    \begin{itemize}[itemsep=0pt,topsep=1pt,leftmargin=12pt]
        \item \textbf{\finetune}: At each cycle, the recommender trained till the last task is trained with the data from the current task.
        \item \textbf{\dropout} \cite{mirzadeh2020dropout}: Dropout \cite{hinton2012improving} is recently found by \cite{mirzadeh2020dropout} that it effectively alleviates catastrophic forgetting. Based on \finetune, we applied dropout to every self-attention and feed-forward layer.
        \item \textbf{\ewc} \cite{kirkpatrick2017overcoming}: It is a well-known method to alleviate forgetting by regularizing parameters important to previous data estimated by the diagonal of a Fisher information matrix computed w.r.t. exemplars.
        %\item Exemplar: Exemplar replay is also a popular method in continue learning. However, the dataset here is quite different from the traditional dataset in the continue learning setting. We introduce our various degree mechanism here to tune the importance of exemplars. We will discuss it in detail in Section \ref{subsec:exemplar}.
\item  \textbf{\ADER} (c.f. Algorithm \ref{alg:UpdateModel}): The  proposed method using adaptively distilled exemplars in each cycle with dropout.
        \item \textbf{\joint}: In each cycle, the recommender is trained (with dropout) using data from the current and \textit{all} historical cycles. This is a common performance ``upper bound'' for continual learning. 
    \end{itemize}
 
The above methods are applied on top of the state-of-the-art base SR recommender SASRec \cite{kang2018self} using 150 hidden units and 2 stacked self-attention blocks. During continual training, we set the batch size to be 256 on DIGINETICA and 512 on YOOCHOOSE. We use Adam optimizer with a learning rate of 5e-4. A total of 100 epochs are trained, and early stop is applied if validation performance (Recall@20) does not improve for 5 consecutive epochs. 
Other hyper-parameters are tuned to maximize Recall@20.
The dropout rate of \dropout, \ADER, and \joint is set to 0.3; 30,000 exemplars are used by default for \ewc\ and \ADER; $\lambda_{base}$ of \ADER\ is set to 0.8 on DIGINETICA and 1.0 on YOOCHOOSE.

\vspace{-0.05in}
    \subsection{Overall Results on Two Datasets}
    \input{table/exp_results}

\begin{figure}
        \centering
        \includegraphics[width=\linewidth]{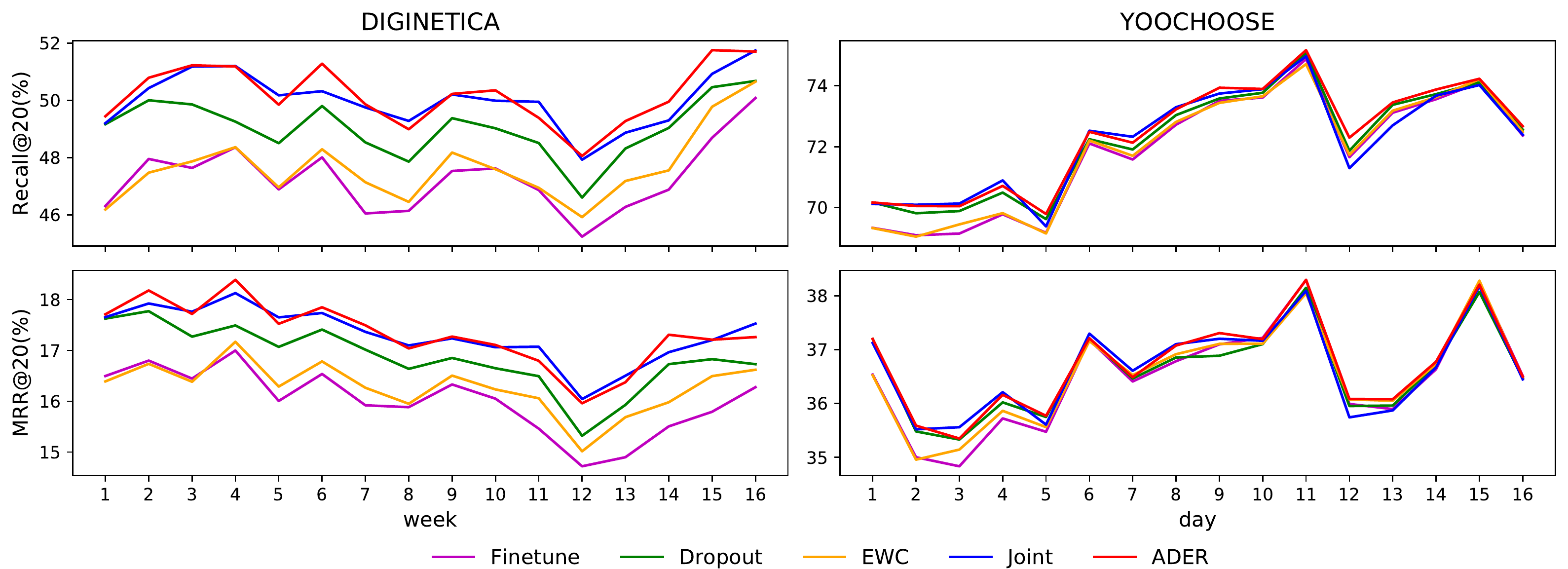}
        \vspace{-0.23in} \caption{Disentangled Recall@20 (Top) and MRR@20 (Bottom) at each continual learning update cycle on two datasets.} \vspace{-0.05in}
        \label{fig:exp}
    \end{figure}

Results averaged over 16 update cycles are presented in Table \ref{table:overall}, and several interesting observations can be noted:
\begin{itemize}[itemsep=0pt,topsep=1pt,leftmargin=12pt]
\item \finetune\ already works reasonably well, especially on the less dynamic YOOCHOOSE dataset. The performance gap between \finetune\ and \joint\ is less significant than typical continual learning setups \cite{rebuffi2017icarl,li2017learning,wu2019large,hou2019learning}. The reason is that catastrophic forgetting is not severe since old items can frequently \textit{reappear} in recommendation tasks. 
\item \ewc\ only outperforms \finetune\ marginally, and it performs worse than \dropout.
\item \dropout\ is effective, and it notably outperforms \finetune, especially on the more dynamic DIGINETICA dataset.
\item \ADER\ significantly outperforms other methods, and the improvement margin over other methods is larger on the more dynamic DIGINETICA dataset. Furthermore, it even outperforms \joint. This result empirically reveals that \ADER\ is a promising solution for the continual recommendation setting by effectively preserving user preference patterns learned before.
\end{itemize}
Detailed disentangled performance at each update cycle is plotted in Figure \ref{fig:exp}. We can see that the advantage of \ADER\ is significant on the more dynamic DIGINETICA dataset. On the less dynamic YOOCHOOSE dataset, the gain of \ADER\ mainly comes from the more dynamic starting cycles with relatively more actions on new items. At later stable cycles with few new items, different methods show comparable performance, including the vanilla \finetune.

\vspace{-0.05in}
\subsection{In-depth Analysis}

In following experiments, we conducted an in-depth analysis of the results on the more dynamic DIGINETICA dataset.

\vspace{-0.05in}
    \subsubsection{\textbf{Different number of Exemplars}}

We studied the effect of a varying number of exemplars for \ADER.
Besides using 30k exemplars, we tested using only 10k/20k exemplars, and results are shown in Table \ref{table:mem_size}. We can see that the performance of \ADER\ only drops marginally as exemplar size decreases from 30k to 10k. This result reveals that \ADER\ is insensitive to the number of exemplars, and it works reasonably well with smaller number of exemplars.

\vspace{-0.05in}
\subsubsection{\textbf{Ablation Study}}

In this experiment, we compared \ADER\ to several simplified versions to justify our design choices. (i). \emph{ER$_{herding}$}: A vanilla exemplar replay different from \ADER\ by using a regular L$_{CE}$, rather than L$_{KD}$, on exemplars. (ii). \emph{ER$_{random}$}: It differs from \emph{ER$_{herding}$} by selecting exemplars of an item at random.
(iii). \emph{ER$_{loss}$}: It differs from \emph{ER$_{herding}$} by selecting exemplars of an item with smallest $L_{CE}$.
(iv). \emph{ADER$_{equal}$}: This version differs from \ADER\ by selecting equal number of exemplars for each item, that is, the assumption that more frequent items should be stored more is removed.
(v). \emph{ADER$_{fix}$}: This version differs from \ADER\ by not using the adaptive $\lambda_t$ in Eq. (\ref{eq:ader_loss}), but a fixed $\lambda$.

Comparison results are presented in Table \ref{table:ablation}, and several observations can be noted: (1). Herding is effective to selected exemplars, indicated by the better performance of \emph{ER$_{herding}$} over \emph{ER$_{random}$} and \emph{ER$_{loss}$}. (2). The distillation loss in Eq. (\ref{eq:distill}) is helpful, indicated by the better performance of three versions of \ADER\ over three vanilla \emph{ER} methods. (3). Selecting exemplars proportional to item frequency is helpful, indicated by the better performance of \ADER\ over \emph{ADER$_{equal}$}. (4). The adaptive $\lambda_t$ in Eq. (\ref{eq:distill}) is helpful, , indicated by the better performance of \ADER\ over \emph{ADER$_{fix}$ }.

\input{table/ablation}

%% file: table/dataset.tex
\vspace{-0.1in}
\begin{table}[t]
  \centering
%   (The \%new item means the proportion of actions on new items among all actions)}
    \begin{tabular}{|c|c|c c c c c c c c c|}
    \hline
        \multirow{6}{*}{\rotatebox{90}{\textbf{DIGINETICA}}} 
        & week 
        & 0 & 1 & 2 & 3 & 4 & 5 & 6 & 7 & 8 \\
        
        \cline{2-11} 
        & total actions 
        & 70,739 & 37,586 & 31,089 & 32,687 & 30,419 & 57,913 
        & 52,225 & 57,100 & 69,042 \\
        
%         \cline{2-11} 
%         & \# new item 
%         & 18,569 & 4,123 & 2,614 & 2,330 & 1,983 & 2,966 
%         & 2,011 & 2,095 & 1,810 \\
        
        \cline{2-11} 
        & new actions & 100.00\% & 18.25\% & 13.26\% & 11.29\% & 10.12\% & 9.08\% & 6.64\% & 6.35\% & 5.42\% \\
        
        \cline{2-11} 
        & week 
        & 9 & 10 & 11 & 12 & 13 & 14 & 15 & 16 & \textbf{Total} \\
        
        \cline{2-11} 
        & total actions  
        & 82,834 & 82,935 & 50,037 & 63,133 & 70,050 & 71,670 
        & 56,959 & 77,065 & 993,483 \\
        
%         \cline{2-11}           
%         & \# new item 
%         & 1,634 & 1,062 & 671 & 501 & 457 & 197 & 82 & 31 & 43,136 \\
        
        \cline{2-11}           
        & new actions  & 5.22\% & 3.02\% & 3.01\% & 1.78\% & 1.83\% & 0.78\% & 0.45\% & 0.27\% & / \\
        \hline
        
        \multirow{6}{*}{\rotatebox{90}{\textbf{YOOCHOOSE}}} 
        
        % & week 
        & day
        & 0 & 1 & 2 & 3 & 4 & 5 & 6 & 7 & 8 \\
        
        \cline{2-11} 
        & total actions 
        % & 12,970,670 & 1,392,230 & 966,103 & 973,953 & 804,018 & 991,293 & 1,044,889 & 1,079,789 & 1,017,711 \\
        & 219,389 & 209,219 & 218,162 & 162,637 & 177,943 & 307,603 & 232,887 & 178,076 & 199,615 \\

%         \cline{2-11} 
%         & \# new item 
%         % & 32425 & 478 & 310 & 482 & 244 & 238 & 242 & 208& 259 \\
%         &12,885 & 2,707 & 1,829 & 1,240 & 979 & 993 & 649 & 721 & 626\\
    
        \cline{2-11} 
        & new actions 
        % & 100\%  & 2.45\% & 3.53\% & 3.87\% & 3.73\% & 4.06\% & 3.93\% & 2.63\% & 4.97\% \\
        & 100.00\% & 3.04\% & 1.74\% & 1.29\% & 0.95\% & 0.57\% & 0.50\% & 1.09\% & 0.74\% \\
        
        \cline{2-11} 
        % & week 
        & day
        & 9 & 10 & 11 & 12 & 13 & 14 & 15 & 16 & \textbf{Total} \\
        
        \cline{2-11} 
        & total actions 
        % & 1,397,835 & 1,540,485 & 1,522,247 & 1,389,485 & 1,217,335 & 1,444,995 & 1,205,134 & 750,333 & 31,708,505 \\
        & 179,889 & 123,750 & 153,565 & 300,830 & 259,673 & 187,348 & 154,316 & 105,676 & 3,370,578
        \\
        
%         \cline{2-11} 
%         & \# new item 
%         % & 233    & 376    & 352    & 716    & 381    & 206    & 180    & 156    & 37,486 \\
%         & 622 & 377 & 447 & 620 & 491 & 312 & 252 & 208 & 25,958\\

        \cline{2-11} 
        & new actions 
        % & 4.92\% & 4.37\% & 2.60\% & 9.09\% & 6.10\% & 6.32\% & 2.93\% & 7.34\% & / \\
        & 0.81\% & 1.08\% & 0.56\% & 0.56\% & 0.29\% & 0.41\% & 0.38\% & 0.35\% & / \\

    \hline
    \end{tabular}%
     \caption{Statistics of the two datasets; ``new actions'' indicate the percentage of actions on new items in this update cycle; week/day 0 is only used for training, while week/day 16 is only used for testing.} \vspace{-0.3in}
  \label{tab:dataset}%
\end{table}%

%% file: table/exp_results.tex
\begin{table}[!t]
  \centering
    \begin{tabular}{|l|c|c|c|c|c|c|c|c|c|c|c|c}
     \hline
    & \multicolumn{5}{c|}{\textbf{DIGINETICA}} & \multicolumn{5}{c|}{\textbf{YOOCHOOSE}} \\
    \hline   
    & \finetune & \dropout & \ewc & \joint & \ADER 
    & \finetune & \dropout & \ewc & \joint & \ADER\\
    \hline
    Recall@20 
    & 47.28\% & 49.07\% & 47.66\% & 50.03\% & \textbf{50.21}\% 
    & 71.86\% & 72.20\% & 71.91\% & 72.22\% & \textbf{72.38\%}\\
    \hline
      Recall@10 
    & 35.00\% & 36.53\% & 35.48\% & 37.27\% & \textbf{37.52}\% 
    & 63.82\% & 64.15\% & 63.89\% & 64.16\% & \textbf{64.41\%} \\
    \hline
    MRR@20 
    & 16.01\% & 16.86\% & 16.28\% & 17.31\% & \textbf{17.32}\% 
    & 36.49\% & 36.60\% & 36.53\% & 36.65\% & \textbf{36.71\%} \\
    \hline
    MRR@10 
    & 15.16\% & 16.00\% & 15.44\% & 16.43\% & \textbf{16.45}\% 
    & 35.92\% & 36.03\% & 35.97\% & 36.08\% & \textbf{36.14\%} \\
    \hline
    \end{tabular}%
      \caption{Performance averaged over 16 continual update cycles on two datasets.} \vspace{-0.35in}
  \label{table:overall}%
\end{table}%

%% file: table/ablation.tex
\begin{figure}[!t]
\noindent
\begin{minipage}{.32\linewidth}
\setlength{\tabcolsep}{3pt}
  \begin{tabular}{|l| c | c |c | }
    \hline
    & 10k & 20k  & 30k   \\
    \hline
    Recall@20  & 49.59\% & 50.05\% & 50.21\%\\
    \hline
    Recall@10 & 36.92\% & 37.40\% & 37.52\% \\
    \hline
    MRR@20 & 17.04\% & 17.29\% & 17.32\% \\
    \hline
   MRR@10 & 16.17\% & 16.42\% &  16.45\% \\
    \hline
    \end{tabular}%
   \captionof{table}{Different exemplar sizes for \ADER.} \vspace{-0.3in}
   \label{table:mem_size}
\end{minipage} \hfill
\begin{minipage}{.65\linewidth}
   \centering
    \setlength{\tabcolsep}{3pt}
   \begin{tabular}{|l| c| c|c|c|c|c| } 
    \hline
    & \emph{ER$_{random}$} & \emph{ER$_{loss}$} & \emph{ER$_{herding}$} & \emph{ADER$_{equal}$} & \emph{ADER$_{fix}$ } & \ADER  \\
    \hline
    Recall@20 &  49.14\% & 49.31\% & 49.34\% & 49.92\% &50.09\% & 50.21\% \\
    \hline
    Recall@10 &  36.61\% & 36.65\% & 36.78\% & 37.21\% & 37.41\% & 37.52\% \\
    \hline
    MRR@20 &  16.79\% &16.90\% & 16.85\% & 17.23\% &17.29\% & 17.32\% \\
    \hline
   MRR@10 &  15.92\% & 16.02\% & 16.98\% & 16.35\% & 16.41\% & 16.45\% \\
    \hline
    \end{tabular}%
   \captionof{table}{Ablation study for \ADER.} \vspace{-0.3in}
   \label{table:ablation}
\end{minipage}

\end{figure}

% \begin{table}[!t]
% \centering
%     \setlength{\tabcolsep}{3pt}
% \begin{subtable}[t]{.65\textwidth}
%     \begin{tabular}{|l| c| c|c|c|c|c| }
%     \hline
%     & ER$_{random}$ & ER$_{loss}$ & ER$_{herding}$ & \emph{ADER-equal} & \emph{ADER-fix} & \ADER  \\
  
%     \hline
%     Recall@20 &  49.14\% & 49.31\% & 49.44\% & 49.92\% & 50.20\% & 50.21\% \\
%     \hline
%     MRR@20 &  16.79\% &16.90\% & 16.95\% & 17.23\% & 17.34\% & 17.32\% \\
%     \hline
%    Recall@10 &  36.61\% & 36.65\% & 36.88\% & 37.21\% & 37.53\% & 37.52\% \\
%     \hline
%    MRR@10 &  15.92\% & 16.02\% & 16.08\% & 16.35\% & 16.46\% & 16.45\% \\
%     \hline
%     \end{tabular}%
%     \caption{Ablation study for \ADER}
% \end{subtable} \quad
% \begin{subtable}[t]{.32\textwidth}
%   \centering
%     \begin{tabular}{|l| c | c |c | }
%     \hline
%     & 10k & 20k  & 30k   \\
%     \hline
%     Recall@20  & 49.59\% & 50.05\% & 50.21\%\\
%     \hline
%     MRR@20 & 17.04\% & 17.29\% & 17.32\% \\
%     \hline
%    Recall@10 & 36.92\% & 37.40\% & 37.52\% \\
%     \hline
%    MRR@10 & 16.17\% & 16.42\% &  16.45\% \\
%     \hline
%     \end{tabular}%
%     \caption{Varying \# of exemplar sizes for \ADER}
% \end{subtable}
% \caption{Detailed analysis for \ADER} \vspace{-0.2in}
% \label{tab:analysis}%
% \end{table}%

%% file: text/conclusion.tex
\section{Conclusion}

In this paper, we studied the practical and realistic continual learning setting for session-based recommendation tasks. To prevent the recommender forgetting user preferences learned before, we propose \ADER\ by replaying carefully chosen exemplars from previous cycles and an adaptive distillation loss. Experiment results on two widely used datasets empirically demonstrate the promising performance of \ADER. Our work may inspire researchers working from similar continual learning perspective for building more robust and scalable recommenders.

%% file: main.bbl

\begin{thebibliography}{39}

%%% the Web address.

\ifx \showCODEN    \undefined \def \showCODEN     #1{\unskip}     \fi
\ifx \showDOI      \undefined \def \showDOI       #1{#1}\fi
\ifx \showISBNx    \undefined \def \showISBNx     #1{\unskip}     \fi
\ifx \showISBNxiii \undefined \def \showISBNxiii  #1{\unskip}     \fi
\ifx \showISSN     \undefined \def \showISSN      #1{\unskip}     \fi
\ifx \showLCCN     \undefined \def \showLCCN      #1{\unskip}     \fi
\ifx \shownote     \undefined \def \shownote      #1{#1}          \fi
\ifx \showarticletitle \undefined \def \showarticletitle #1{#1}   \fi
\ifx \showURL      \undefined \def \showURL       {\relax}        \fi
% invisible to TeX
\providecommand\bibfield[2]{#2}
\providecommand\bibinfo[2]{#2}
\providecommand\natexlab[1]{#1}
\providecommand\showeprint[2][]{arXiv:#2}

\bibitem[\protect\citeauthoryear{Aljundi, Babiloni, Elhoseiny, Rohrbach, and
  Tuytelaars}{Aljundi et~al\mbox{.}}{2018}]%
        {aljundi2018memory}
\bibfield{author}{\bibinfo{person}{Rahaf Aljundi}, \bibinfo{person}{Francesca
  Babiloni}, \bibinfo{person}{Mohamed Elhoseiny}, \bibinfo{person}{Marcus
  Rohrbach}, {and} \bibinfo{person}{Tinne Tuytelaars}.}
  \bibinfo{year}{2018}\natexlab{}.
\newblock \showarticletitle{Memory aware synapses: Learning what (not) to
  forget}. In \bibinfo{booktitle}{\emph{Proceedings of the European Conference
  on Computer Vision (ECCV)}}. \bibinfo{pages}{139--154}.
\newblock


\bibitem[\protect\citeauthoryear{Andrychowicz, Wolski, Ray, Schneider, Fong,
  Welinder, McGrew, Tobin, Abbeel, and Zaremba}{Andrychowicz
  et~al\mbox{.}}{2017}]%
        {andrychowicz2017hindsight}
\bibfield{author}{\bibinfo{person}{Marcin Andrychowicz}, \bibinfo{person}{Filip
  Wolski}, \bibinfo{person}{Alex Ray}, \bibinfo{person}{Jonas Schneider},
  \bibinfo{person}{Rachel Fong}, \bibinfo{person}{Peter Welinder},
  \bibinfo{person}{Bob McGrew}, \bibinfo{person}{Josh Tobin},
  \bibinfo{person}{OpenAI~Pieter Abbeel}, {and} \bibinfo{person}{Wojciech
  Zaremba}.} \bibinfo{year}{2017}\natexlab{}.
\newblock \showarticletitle{Hindsight experience replay}. In
  \bibinfo{booktitle}{\emph{Advances in neural information processing
  systems}}. \bibinfo{pages}{5048--5058}.
\newblock


\bibitem[\protect\citeauthoryear{Castro, Mar{\'\i}n-Jim{\'e}nez, Guil, Schmid,
  and Alahari}{Castro et~al\mbox{.}}{2018}]%
        {castro2018end}
\bibfield{author}{\bibinfo{person}{Francisco~M Castro},
  \bibinfo{person}{Manuel~J Mar{\'\i}n-Jim{\'e}nez},
  \bibinfo{person}{Nicol{\'a}s Guil}, \bibinfo{person}{Cordelia Schmid}, {and}
  \bibinfo{person}{Karteek Alahari}.} \bibinfo{year}{2018}\natexlab{}.
\newblock \showarticletitle{End-to-end incremental learning}. In
  \bibinfo{booktitle}{\emph{Proceedings of the European Conference on Computer
  Vision (ECCV)}}. \bibinfo{pages}{233--248}.
\newblock


\bibitem[\protect\citeauthoryear{Chaudhry, Rohrbach, Elhoseiny, Ajanthan,
  Dokania, Torr, and Ranzato}{Chaudhry et~al\mbox{.}}{2019}]%
        {chaudhry2019continual}
\bibfield{author}{\bibinfo{person}{Arslan Chaudhry}, \bibinfo{person}{Marcus
  Rohrbach}, \bibinfo{person}{Mohamed Elhoseiny},
  \bibinfo{person}{Thalaiyasingam Ajanthan}, \bibinfo{person}{Puneet~K
  Dokania}, \bibinfo{person}{Philip~HS Torr}, {and}
  \bibinfo{person}{Marc'Aurelio Ranzato}.} \bibinfo{year}{2019}\natexlab{}.
\newblock \showarticletitle{Continual Learning with Tiny Episodic Memories}.
\newblock \bibinfo{journal}{\emph{arXiv preprint arXiv:1902.10486}}
  (\bibinfo{year}{2019}).
\newblock


\bibitem[\protect\citeauthoryear{Devlin, Chang, Lee, and Toutanova}{Devlin
  et~al\mbox{.}}{2018}]%
        {devlin2018bert}
\bibfield{author}{\bibinfo{person}{Jacob Devlin}, \bibinfo{person}{Ming-Wei
  Chang}, \bibinfo{person}{Kenton Lee}, {and} \bibinfo{person}{Kristina
  Toutanova}.} \bibinfo{year}{2018}\natexlab{}.
\newblock \showarticletitle{Bert: Pre-training of deep bidirectional
  transformers for language understanding}.
\newblock \bibinfo{journal}{\emph{arXiv preprint arXiv:1810.04805}}
  (\bibinfo{year}{2018}).
\newblock


\bibitem[\protect\citeauthoryear{French}{French}{1999}]%
        {french1999catastrophic}
\bibfield{author}{\bibinfo{person}{Robert~M French}.}
  \bibinfo{year}{1999}\natexlab{}.
\newblock \showarticletitle{Catastrophic forgetting in connectionist networks}.
\newblock \bibinfo{journal}{\emph{Trends in Cognitive Sciences}}
  (\bibinfo{year}{1999}), \bibinfo{pages}{128--135}.
\newblock


\bibitem[\protect\citeauthoryear{Garcin, Dimitrakakis, and Faltings}{Garcin
  et~al\mbox{.}}{2013}]%
        {garcin2013personalized}
\bibfield{author}{\bibinfo{person}{Florent Garcin}, \bibinfo{person}{Christos
  Dimitrakakis}, {and} \bibinfo{person}{Boi Faltings}.}
  \bibinfo{year}{2013}\natexlab{}.
\newblock \showarticletitle{Personalized news recommendation with context
  trees}. In \bibinfo{booktitle}{\emph{RecSys}}. ACM,
  \bibinfo{pages}{105--112}.
\newblock


\bibitem[\protect\citeauthoryear{Garg, Gupta, Malhotra, Vig, and Shroff}{Garg
  et~al\mbox{.}}{2019}]%
        {garg2019sequence}
\bibfield{author}{\bibinfo{person}{Diksha Garg}, \bibinfo{person}{Priyanka
  Gupta}, \bibinfo{person}{Pankaj Malhotra}, \bibinfo{person}{Lovekesh Vig},
  {and} \bibinfo{person}{Gautam Shroff}.} \bibinfo{year}{2019}\natexlab{}.
\newblock \showarticletitle{Sequence and time aware neighborhood for
  session-based recommendations: Stan}. In \bibinfo{booktitle}{\emph{SIGIR}}.
  \bibinfo{pages}{1069--1072}.
\newblock


\bibitem[\protect\citeauthoryear{Guo, Yin, Wang, Chen, Zhou, and Quoc
  Viet~Hung}{Guo et~al\mbox{.}}{2019}]%
        {guo2019streaming}
\bibfield{author}{\bibinfo{person}{Lei Guo}, \bibinfo{person}{Hongzhi Yin},
  \bibinfo{person}{Qinyong Wang}, \bibinfo{person}{Tong Chen},
  \bibinfo{person}{Alexander Zhou}, {and} \bibinfo{person}{Nguyen Quoc
  Viet~Hung}.} \bibinfo{year}{2019}\natexlab{}.
\newblock \showarticletitle{Streaming session-based recommendation}. In
  \bibinfo{booktitle}{\emph{Proceedings of the 25th ACM SIGKDD International
  Conference on Knowledge Discovery \& Data Mining}}.
  \bibinfo{pages}{1569--1577}.
\newblock


\bibitem[\protect\citeauthoryear{Hidasi and Karatzoglou}{Hidasi and
  Karatzoglou}{2018}]%
        {hidasi2018recurrent}
\bibfield{author}{\bibinfo{person}{Bal{\'a}zs Hidasi} {and}
  \bibinfo{person}{Alexandros Karatzoglou}.} \bibinfo{year}{2018}\natexlab{}.
\newblock \showarticletitle{Recurrent neural networks with top-k gains for
  session-based recommendations}. In \bibinfo{booktitle}{\emph{Proceedings of
  the 27th ACM International Conference on Information and Knowledge
  Management}}. \bibinfo{pages}{843--852}.
\newblock


\bibitem[\protect\citeauthoryear{Hidasi, Karatzoglou, Baltrunas, and
  Tikk}{Hidasi et~al\mbox{.}}{2016}]%
        {hidasi2015session}
\bibfield{author}{\bibinfo{person}{Bal{\'a}zs Hidasi},
  \bibinfo{person}{Alexandros Karatzoglou}, \bibinfo{person}{Linas Baltrunas},
  {and} \bibinfo{person}{Domonkos Tikk}.} \bibinfo{year}{2016}\natexlab{}.
\newblock \showarticletitle{Session-based recommendations with recurrent neural
  networks}. In \bibinfo{booktitle}{\emph{ICLR}}.
\newblock


\bibitem[\protect\citeauthoryear{Hinton, Vinyals, and Dean}{Hinton
  et~al\mbox{.}}{2015}]%
        {hinton2015distilling}
\bibfield{author}{\bibinfo{person}{Geoffrey Hinton}, \bibinfo{person}{Oriol
  Vinyals}, {and} \bibinfo{person}{Jeff Dean}.}
  \bibinfo{year}{2015}\natexlab{}.
\newblock \showarticletitle{Distilling the knowledge in a neural network}.
\newblock \bibinfo{journal}{\emph{arXiv preprint arXiv:1503.02531}}
  (\bibinfo{year}{2015}).
\newblock


\bibitem[\protect\citeauthoryear{Hinton, Srivastava, Krizhevsky, Sutskever, and
  Salakhutdinov}{Hinton et~al\mbox{.}}{2012}]%
        {hinton2012improving}
\bibfield{author}{\bibinfo{person}{Geoffrey~E Hinton}, \bibinfo{person}{Nitish
  Srivastava}, \bibinfo{person}{Alex Krizhevsky}, \bibinfo{person}{Ilya
  Sutskever}, {and} \bibinfo{person}{Ruslan~R Salakhutdinov}.}
  \bibinfo{year}{2012}\natexlab{}.
\newblock \showarticletitle{Improving neural networks by preventing
  co-adaptation of feature detectors}.
\newblock \bibinfo{journal}{\emph{arXiv preprint arXiv:1207.0580}}
  (\bibinfo{year}{2012}).
\newblock


\bibitem[\protect\citeauthoryear{Hou, Pan, Loy, Wang, and Lin}{Hou
  et~al\mbox{.}}{2019}]%
        {hou2019learning}
\bibfield{author}{\bibinfo{person}{Saihui Hou}, \bibinfo{person}{Xinyu Pan},
  \bibinfo{person}{Chen~Change Loy}, \bibinfo{person}{Zilei Wang}, {and}
  \bibinfo{person}{Dahua Lin}.} \bibinfo{year}{2019}\natexlab{}.
\newblock \showarticletitle{Learning a unified classifier incrementally via
  rebalancing}. In \bibinfo{booktitle}{\emph{Proceedings of the IEEE Conference
  on Computer Vision and Pattern Recognition}}. \bibinfo{pages}{831--839}.
\newblock


\bibitem[\protect\citeauthoryear{Jannach and Ludewig}{Jannach and
  Ludewig}{2017}]%
        {jannach2017recurrent}
\bibfield{author}{\bibinfo{person}{Dietmar Jannach} {and}
  \bibinfo{person}{Malte Ludewig}.} \bibinfo{year}{2017}\natexlab{}.
\newblock \showarticletitle{When recurrent neural networks meet the
  neighborhood for session-based recommendation}. In
  \bibinfo{booktitle}{\emph{RecSys}}. ACM, \bibinfo{pages}{306--310}.
\newblock


\bibitem[\protect\citeauthoryear{Kang and McAuley}{Kang and McAuley}{2018}]%
        {kang2018self}
\bibfield{author}{\bibinfo{person}{Wang-Cheng Kang} {and}
  \bibinfo{person}{Julian McAuley}.} \bibinfo{year}{2018}\natexlab{}.
\newblock \showarticletitle{Self-attentive sequential recommendation}. In
  \bibinfo{booktitle}{\emph{2018 IEEE International Conference on Data Mining
  (ICDM)}}. IEEE, \bibinfo{pages}{197--206}.
\newblock


\bibitem[\protect\citeauthoryear{Kirkpatrick, Pascanu, Rabinowitz, Veness,
  Desjardins, Rusu, Milan, Quan, Ramalho, Grabska-Barwinska,
  et~al\mbox{.}}{Kirkpatrick et~al\mbox{.}}{2017}]%
        {kirkpatrick2017overcoming}
\bibfield{author}{\bibinfo{person}{James Kirkpatrick}, \bibinfo{person}{Razvan
  Pascanu}, \bibinfo{person}{Neil Rabinowitz}, \bibinfo{person}{Joel Veness},
  \bibinfo{person}{Guillaume Desjardins}, \bibinfo{person}{Andrei~A Rusu},
  \bibinfo{person}{Kieran Milan}, \bibinfo{person}{John Quan},
  \bibinfo{person}{Tiago Ramalho}, \bibinfo{person}{Agnieszka
  Grabska-Barwinska}, {et~al\mbox{.}}} \bibinfo{year}{2017}\natexlab{}.
\newblock \showarticletitle{Overcoming catastrophic forgetting in neural
  networks}.
\newblock \bibinfo{journal}{\emph{Proceedings of the national academy of
  sciences}} \bibinfo{volume}{114}, \bibinfo{number}{13}
  (\bibinfo{year}{2017}), \bibinfo{pages}{3521--3526}.
\newblock


\bibitem[\protect\citeauthoryear{Li, Ren, Chen, Ren, Lian, and Ma}{Li
  et~al\mbox{.}}{2017}]%
        {li2017neural}
\bibfield{author}{\bibinfo{person}{Jing Li}, \bibinfo{person}{Pengjie Ren},
  \bibinfo{person}{Zhumin Chen}, \bibinfo{person}{Zhaochun Ren},
  \bibinfo{person}{Tao Lian}, {and} \bibinfo{person}{Jun Ma}.}
  \bibinfo{year}{2017}\natexlab{}.
\newblock \showarticletitle{Neural attentive session-based recommendation}. In
  \bibinfo{booktitle}{\emph{Proceedings of the 2017 ACM on Conference on
  Information and Knowledge Management}}. \bibinfo{pages}{1419--1428}.
\newblock


\bibitem[\protect\citeauthoryear{Li and Hoiem}{Li and Hoiem}{2017}]%
        {li2017learning}
\bibfield{author}{\bibinfo{person}{Zhizhong Li} {and} \bibinfo{person}{Derek
  Hoiem}.} \bibinfo{year}{2017}\natexlab{}.
\newblock \showarticletitle{Learning without forgetting}.
\newblock \bibinfo{journal}{\emph{IEEE transactions on pattern analysis and
  machine intelligence}} \bibinfo{volume}{40}, \bibinfo{number}{12}
  (\bibinfo{year}{2017}), \bibinfo{pages}{2935--2947}.
\newblock


\bibitem[\protect\citeauthoryear{Liu, Zeng, Mokhosi, and Zhang}{Liu
  et~al\mbox{.}}{2018}]%
        {liu2018stamp}
\bibfield{author}{\bibinfo{person}{Qiao Liu}, \bibinfo{person}{Yifu Zeng},
  \bibinfo{person}{Refuoe Mokhosi}, {and} \bibinfo{person}{Haibin Zhang}.}
  \bibinfo{year}{2018}\natexlab{}.
\newblock \showarticletitle{STAMP: short-term attention/memory priority model
  for session-based recommendation}. In \bibinfo{booktitle}{\emph{Proceedings
  of the 24th ACM SIGKDD International Conference on Knowledge Discovery \&
  Data Mining}}. \bibinfo{pages}{1831--1839}.
\newblock


\bibitem[\protect\citeauthoryear{Ludewig and Jannach}{Ludewig and
  Jannach}{2018}]%
        {ludewig2018evaluation}
\bibfield{author}{\bibinfo{person}{Malte Ludewig} {and}
  \bibinfo{person}{Dietmar Jannach}.} \bibinfo{year}{2018}\natexlab{}.
\newblock \showarticletitle{Evaluation of session-based recommendation
  algorithms}.
\newblock \bibinfo{journal}{\emph{User Modeling and User-Adapted Interaction}}
  \bibinfo{volume}{28}, \bibinfo{number}{4-5} (\bibinfo{year}{2018}),
  \bibinfo{pages}{331--390}.
\newblock


\bibitem[\protect\citeauthoryear{Maltoni and Lomonaco}{Maltoni and
  Lomonaco}{2019}]%
        {maltoni2019continuous}
\bibfield{author}{\bibinfo{person}{Davide Maltoni} {and}
  \bibinfo{person}{Vincenzo Lomonaco}.} \bibinfo{year}{2019}\natexlab{}.
\newblock \showarticletitle{Continuous learning in single-incremental-task
  scenarios}.
\newblock \bibinfo{journal}{\emph{Neural Networks}}  \bibinfo{volume}{116}
  (\bibinfo{year}{2019}), \bibinfo{pages}{56--73}.
\newblock


\bibitem[\protect\citeauthoryear{McCloskey and Cohen}{McCloskey and
  Cohen}{1989}]%
        {mccloskey1989catastrophic}
\bibfield{author}{\bibinfo{person}{Michael McCloskey} {and}
  \bibinfo{person}{Neal~J Cohen}.} \bibinfo{year}{1989}\natexlab{}.
\newblock \showarticletitle{Catastrophic interference in connectionist
  networks: The sequential learning problem}.
\newblock In \bibinfo{booktitle}{\emph{Psychology of learning and motivation}}.
  Vol.~\bibinfo{volume}{24}. \bibinfo{publisher}{Elsevier},
  \bibinfo{pages}{109--165}.
\newblock


\bibitem[\protect\citeauthoryear{Mi and Faltings}{Mi and Faltings}{2016}]%
        {mi2016adaptive}
\bibfield{author}{\bibinfo{person}{Fei Mi} {and} \bibinfo{person}{Boi
  Faltings}.} \bibinfo{year}{2016}\natexlab{}.
\newblock \showarticletitle{Adaptive Sequential Recommendation Using Context
  Trees.}. In \bibinfo{booktitle}{\emph{IJCAI}}. \bibinfo{pages}{4018--4019}.
\newblock


\bibitem[\protect\citeauthoryear{Mi and Faltings}{Mi and Faltings}{2017}]%
        {mi2017adaptive}
\bibfield{author}{\bibinfo{person}{Fei Mi} {and} \bibinfo{person}{Boi
  Faltings}.} \bibinfo{year}{2017}\natexlab{}.
\newblock \showarticletitle{Adaptive sequential recommendation for discussion
  forums on MOOCs using context trees}. In
  \bibinfo{booktitle}{\emph{Proceedings of the 10th international conference on
  educational data mining}}.
\newblock


\bibitem[\protect\citeauthoryear{Mi and Faltings}{Mi and Faltings}{2018}]%
        {mi2018context}
\bibfield{author}{\bibinfo{person}{Fei Mi} {and} \bibinfo{person}{Boi
  Faltings}.} \bibinfo{year}{2018}\natexlab{}.
\newblock \showarticletitle{Context Tree for Adaptive Session-based
  Recommendation}.
\newblock \bibinfo{journal}{\emph{arXiv preprint arXiv:1806.03733}}
  (\bibinfo{year}{2018}).
\newblock


\bibitem[\protect\citeauthoryear{Mi and Faltings}{Mi and Faltings}{2020}]%
        {mi2020memory}
\bibfield{author}{\bibinfo{person}{Fei Mi} {and} \bibinfo{person}{Boi
  Faltings}.} \bibinfo{year}{2020}\natexlab{}.
\newblock \showarticletitle{Memory Augmented Neural Model for Incremental
  Session-based Recommendation}.
\newblock \bibinfo{journal}{\emph{arXiv preprint arXiv:2005.01573}}
  (\bibinfo{year}{2020}).
\newblock


\bibitem[\protect\citeauthoryear{Mirzadeh, Farajtabar, and
  Ghasemzadeh}{Mirzadeh et~al\mbox{.}}{2020}]%
        {mirzadeh2020dropout}
\bibfield{author}{\bibinfo{person}{Seyed-Iman Mirzadeh},
  \bibinfo{person}{Mehrdad Farajtabar}, {and} \bibinfo{person}{Hassan
  Ghasemzadeh}.} \bibinfo{year}{2020}\natexlab{}.
\newblock \showarticletitle{Dropout as an Implicit Gating Mechanism For
  Continual Learning}.
\newblock \bibinfo{journal}{\emph{arXiv preprint arXiv:2004.11545}}
  (\bibinfo{year}{2020}).
\newblock


\bibitem[\protect\citeauthoryear{Rebuffi, Kolesnikov, Sperl, and
  Lampert}{Rebuffi et~al\mbox{.}}{2017}]%
        {rebuffi2017icarl}
\bibfield{author}{\bibinfo{person}{Sylvestre-Alvise Rebuffi},
  \bibinfo{person}{Alexander Kolesnikov}, \bibinfo{person}{Georg Sperl}, {and}
  \bibinfo{person}{Christoph~H Lampert}.} \bibinfo{year}{2017}\natexlab{}.
\newblock \showarticletitle{icarl: Incremental classifier and representation
  learning}. In \bibinfo{booktitle}{\emph{Proceedings of the IEEE conference on
  Computer Vision and Pattern Recognition}}. \bibinfo{pages}{2001--2010}.
\newblock


\bibitem[\protect\citeauthoryear{Ren, Chen, Li, Ren, Ma, and de~Rijke}{Ren
  et~al\mbox{.}}{2019}]%
        {ren2018repeatnet}
\bibfield{author}{\bibinfo{person}{Pengjie Ren}, \bibinfo{person}{Zhumin Chen},
  \bibinfo{person}{Jing Li}, \bibinfo{person}{Zhaochun Ren},
  \bibinfo{person}{Jun Ma}, {and} \bibinfo{person}{Maarten de Rijke}.}
  \bibinfo{year}{2019}\natexlab{}.
\newblock \showarticletitle{RepeatNet: A Repeat Aware Neural Recommendation
  Machine for Session-based Recommendation}. In
  \bibinfo{booktitle}{\emph{AAAI}}.
\newblock


\bibitem[\protect\citeauthoryear{Rusu, Rabinowitz, Desjardins, Soyer,
  Kirkpatrick, Kavukcuoglu, Pascanu, and Hadsell}{Rusu et~al\mbox{.}}{2016}]%
        {rusu2016progressive}
\bibfield{author}{\bibinfo{person}{Andrei~A Rusu}, \bibinfo{person}{Neil~C
  Rabinowitz}, \bibinfo{person}{Guillaume Desjardins}, \bibinfo{person}{Hubert
  Soyer}, \bibinfo{person}{James Kirkpatrick}, \bibinfo{person}{Koray
  Kavukcuoglu}, \bibinfo{person}{Razvan Pascanu}, {and} \bibinfo{person}{Raia
  Hadsell}.} \bibinfo{year}{2016}\natexlab{}.
\newblock \showarticletitle{Progressive neural networks}.
\newblock \bibinfo{journal}{\emph{arXiv preprint arXiv:1606.04671}}
  (\bibinfo{year}{2016}).
\newblock


\bibitem[\protect\citeauthoryear{Schaul, Quan, Antonoglou, and Silver}{Schaul
  et~al\mbox{.}}{2016}]%
        {schaul2015prioritized}
\bibfield{author}{\bibinfo{person}{Tom Schaul}, \bibinfo{person}{John Quan},
  \bibinfo{person}{Ioannis Antonoglou}, {and} \bibinfo{person}{David Silver}.}
  \bibinfo{year}{2016}\natexlab{}.
\newblock \showarticletitle{Prioritized experience replay}.
\newblock  (\bibinfo{year}{2016}).
\newblock


\bibitem[\protect\citeauthoryear{Sun, Liu, Wu, Pei, Lin, Ou, and Jiang}{Sun
  et~al\mbox{.}}{2019}]%
        {sun2019bert4rec}
\bibfield{author}{\bibinfo{person}{Fei Sun}, \bibinfo{person}{Jun Liu},
  \bibinfo{person}{Jian Wu}, \bibinfo{person}{Changhua Pei},
  \bibinfo{person}{Xiao Lin}, \bibinfo{person}{Wenwu Ou}, {and}
  \bibinfo{person}{Peng Jiang}.} \bibinfo{year}{2019}\natexlab{}.
\newblock \showarticletitle{BERT4Rec: Sequential recommendation with
  bidirectional encoder representations from transformer}. In
  \bibinfo{booktitle}{\emph{CIKM}}. \bibinfo{pages}{1441--1450}.
\newblock


\bibitem[\protect\citeauthoryear{Tan, Xu, and Liu}{Tan et~al\mbox{.}}{2016}]%
        {tan2016improved}
\bibfield{author}{\bibinfo{person}{Yong~Kiam Tan}, \bibinfo{person}{Xinxing
  Xu}, {and} \bibinfo{person}{Yong Liu}.} \bibinfo{year}{2016}\natexlab{}.
\newblock \showarticletitle{Improved recurrent neural networks for
  session-based recommendations}. In \bibinfo{booktitle}{\emph{Proceedings of
  the 1st Workshop on Deep Learning for Recommender Systems}}.
  \bibinfo{pages}{17--22}.
\newblock


\bibitem[\protect\citeauthoryear{Vaswani, Shazeer, Parmar, Uszkoreit, Jones,
  Gomez, Kaiser, and Polosukhin}{Vaswani et~al\mbox{.}}{2017}]%
        {vaswani2017attention}
\bibfield{author}{\bibinfo{person}{Ashish Vaswani}, \bibinfo{person}{Noam
  Shazeer}, \bibinfo{person}{Niki Parmar}, \bibinfo{person}{Jakob Uszkoreit},
  \bibinfo{person}{Llion Jones}, \bibinfo{person}{Aidan~N Gomez},
  \bibinfo{person}{{\L}ukasz Kaiser}, {and} \bibinfo{person}{Illia
  Polosukhin}.} \bibinfo{year}{2017}\natexlab{}.
\newblock \showarticletitle{Attention is all you need}. In
  \bibinfo{booktitle}{\emph{Advances in neural information processing
  systems}}. \bibinfo{pages}{5998--6008}.
\newblock


\bibitem[\protect\citeauthoryear{Welling}{Welling}{2009}]%
        {welling2009herding}
\bibfield{author}{\bibinfo{person}{Max Welling}.}
  \bibinfo{year}{2009}\natexlab{}.
\newblock \showarticletitle{Herding dynamical weights to learn}. In
  \bibinfo{booktitle}{\emph{Proceedings of the 26th Annual International
  Conference on Machine Learning}}. \bibinfo{pages}{1121--1128}.
\newblock


\bibitem[\protect\citeauthoryear{Wu, Chen, Wang, Ye, Liu, Guo, and Fu}{Wu
  et~al\mbox{.}}{2019}]%
        {wu2019large}
\bibfield{author}{\bibinfo{person}{Yue Wu}, \bibinfo{person}{Yinpeng Chen},
  \bibinfo{person}{Lijuan Wang}, \bibinfo{person}{Yuancheng Ye},
  \bibinfo{person}{Zicheng Liu}, \bibinfo{person}{Yandong Guo}, {and}
  \bibinfo{person}{Yun Fu}.} \bibinfo{year}{2019}\natexlab{}.
\newblock \showarticletitle{Large scale incremental learning}. In
  \bibinfo{booktitle}{\emph{Proceedings of the IEEE Conference on Computer
  Vision and Pattern Recognition}}. \bibinfo{pages}{374--382}.
\newblock


\bibitem[\protect\citeauthoryear{Zenke, Poole, and Ganguli}{Zenke
  et~al\mbox{.}}{2017}]%
        {zenke2017continual}
\bibfield{author}{\bibinfo{person}{Friedemann Zenke}, \bibinfo{person}{Ben
  Poole}, {and} \bibinfo{person}{Surya Ganguli}.}
  \bibinfo{year}{2017}\natexlab{}.
\newblock \showarticletitle{Continual learning through synaptic intelligence}.
  In \bibinfo{booktitle}{\emph{Proceedings of the 34th International Conference
  on Machine Learning}}. JMLR. org, \bibinfo{pages}{3987--3995}.
\newblock


\bibitem[\protect\citeauthoryear{Zhao, Xiao, Gan, Zhang, and Xia}{Zhao
  et~al\mbox{.}}{2019}]%
        {zhao2019maintaining}
\bibfield{author}{\bibinfo{person}{Bowen Zhao}, \bibinfo{person}{Xi Xiao},
  \bibinfo{person}{Guojun Gan}, \bibinfo{person}{Bin Zhang}, {and}
  \bibinfo{person}{Shutao Xia}.} \bibinfo{year}{2019}\natexlab{}.
\newblock \showarticletitle{Maintaining Discrimination and Fairness in Class
  Incremental Learning}.
\newblock \bibinfo{journal}{\emph{arXiv preprint arXiv:1911.07053}}
  (\bibinfo{year}{2019}).
\newblock


\end{thebibliography}
